# N-gram Statistical Stemmer for Bangla Corpus

Rabeya Sadia[a], Md. Ataur Rahman[b] and Md. Hanif Seddiqui[a]

[a]Dept. of Computer Science & Engineering, University of Chittagong, Chittagong-4331, Bangladesh
[b]Dept. of Language Science and Technology, University of Saarland, 66125 Saarbrücken, Germany

Email:rabeyasadia8@gmail.com

**Abstract**

Stemming is a process that can be utilized to trim inflected words to stem or root form. It is useful for enhancing the retrieval effectiveness, especially for text search in order to solve the mismatch problems. Previous research on Bangla stemming mostly relied on eliminating multiple suffixes from a solitary word through a recursive rule based procedure to recover progressively applicable relative root. Our proposed system has enhanced the aforementioned exploration by actualizing one of the stemming algorithms called N-gram stemming. By utilizing an affiliation measure called dice coefficient, related sets of words are clustered depending on their character structure. The smallest word in one cluster may be considered as the stem. We additionally analyzed Affinity Propagation clustering algorithms with coefficient similarity as well as with median similarity. Our result indicates N-gram stemming techniques to be effective in general which gave us around 87% accurate clusters.

## 1.0 Introduction

Stemming is important in Natural Language Processing (NLP) that extracts meaningful information from immense sources like big data or the web [1]. Bangla is an exceptionally inflectional language containing several morphological variances on a single stem [2]. Terms with common stems tend to have similar meaning, utilized in information retrieval to defeat the vocabulary mismatch problem. Various methodologies are utilized in stemming in order to decrease a word to its base structure no matter which inflected form is experienced. It is typically done by splitting terms into its constituent main part and a single or a number of suffixes. Conventional rule-based stemmers tailor the phonetic learning of specialists. Differently, statistical stemmers provide language-independent approaches that cluster the related set of words based on different string-similarity measures[4]. It is not essential that the root word should be the lexical root. But if the root word is lexical root then it is simpler to improve the recall that enhances the retrieval effectiveness. To beat this impediment, we propose a stemming algorithm that is capable of clustering lexically related words based on their character structure by utilizing an association measure named dice coefficient. To this end, we also introduce an unsupervised method called Affinity Propagation clustering algorithm to cluster all morphological variations of a word. The morphological variations of the words based on a given large collection of documents are readily available on the web which is NLP collection. Our examinations and evaluations on a large corpus provides 87% accuracy.

The rest of the paper is composed of different sections. Some of the related works are explained in section 2.0. Dataset preparation is presented in section 3.0. Experimental settings and methodology are illustrated in section 4.0. Results and evaluation are demonstrated in section 5.0. A brief description of future work is given in section 6.0 to conclude the work.

## 2.0 Literature Review

Although the English language is enriched with lots of stemmers, the available stemmers in Bangla mostly incorporates predetermined suffix list in order to strip the affixes from a word.

An unsupervised morphological parsing was proposed by Dasgupta and Ng [5] to segment a word into multiple roots by defining the parser into prefix, suffix, and stem, a challenging research problem. They evaluated on a set of 4,110 hand-segmented Bengali words collected from news corpus and the algorithm obtained an F-score of 83%, substantially outperforming *Linguistica*, a standout amongst the most broadly utilized unsupervised morphological parsers, by about 23% in F-score.

Islam et al. (2007) [6] suggested an extreme lightweight stemmer based on a 'longest match'. The authors claimed that the algorithm stripped the suffixes using the predetermined suffix list. They collected 600 root words and 100 suffixes containing 72 suffixes for verbs, 22 for nouns and 8 for adjectives for Bangla language. The proposed stemming algorithm is primarily to deal with inflections.

Majumder et al. (2007) [7] proposed a statistical approach *Yet Another Suffix Stripper* (YASS) that uses clustering techniques based on the string similarity measure. It required no earlier phonetic learning to cluster a lexicon from a content corpus into homogeneous gatherings containing comparative morphological variations of a single root word using Graph-theoretic grouping algorithm.

Hanif et al. (2016) [2] used a Recursive Suffix Stripping to Augment Bangla Stemmer that introduces a recursive process to eliminate multiple suffixes from a single word. The proposed algorithm used conservative, aggressive, and rule-based approaches where an inflectional word is stemmed in all possible ways to identify the final stem. They evaluated on relatively larger text corpora and the algorithm obtained 92% accuracy.

**3.0 Dataset**

We collected our data by scraping article from Bangla Wikipedia[1] and a popular Bangla daily newspaper, The Daily Prothom Alo[2]. The dataset contains a total of 46892 words (Table 1) and after performing preprocessing 10133 unique words are found that is used for training purpose.

Table 1. Dataset

| | |
|---|---|
| Total words | 46892 |
| Unique words | 10133 |

**4.0 Experimental Settings and Methodology**

This section consists of experimental approaches that we took to cluster the related set of words and to form the corresponding root words. At first, preprocessing is performed to convert the raw data into an appropriate form for training. Then by using trainable dataset, we experiment on N-gram stemming approach and affinity propagation clustering algorithm with coefficient similarity and with median similarity. And depending on the model's outcomes the best model is preferred.

*4.1 Preprocessing*

Preprocessing is a combination of several individual steps. To convert the raw data into a trainable form we perform the following steps in preprocessing:
       a) Clean the plain text by removing punctuation's, digits, emoticons etc. except bangla word.
       b) Separate each word in the sentence of plain text.
       c) Perform tokenization.
       d) Remove Bangla words containing just only one character.

---

[1] **https://bn.wikipedia.org/wiki/বাংলা_উইকিপিডিয়া**
[2] **https://www.prothomalo.com/**

e) Make a list of unique words and remove whitespaces.

At first there was a total of 46892 tokens in the raw documents, the preprocessing reduced it to 10133 unique words.

*4.2 N-gram Stemmer*

N-gram stemmer make use of *n* sequence of characters within a word in order to determine the similarity. More specifically, *bi-gram* (n=2) or *tri-gram* (n=3) can be used to calculate association measure between a pair of words based on shared di-gram method [3].

**Similarity Coefficient:** The association measure that is used in the N-gram approach is the dice coefficient[8] because of its simplicity which is defined as

$$S=2C/(A+B)$$

C = common distinct character bi-grams (or tri-grams) of the two words.
A = the number of distinct character bi-grams (or tri-grams) in the 1st word.
B = the number of distinct character bi-grams (or tri-grams) in the 2nd word.

**Coefficients Calculation:** To compute the coefficient similarity between two words, the character bi-grams or tri-grams of each word is generated and stored. The number of co-occurring unique bi-grams or tri-grams are calculated by comparing both of the words. The sum of unique bi-grams or tri-grams in each word is counted for the final computation. Once the similarity coefficient for two words is obtained, the value is compared using a threshold value (0.06). If the value of coefficient similarity is higher than the threshold value then these two words are placed in the same cluster where the smallest word in the cluster is selected as stem. The whole procedure is continued until the empty token list found.

*4.3 Affinity Propagation*

Affinity propagation is a clustering algorithm which is a combination of two variants: a class is used to implement the fit method to be told the clusters on train data and a function is used that takes the train data as input and returns an array of integer labels for different clusters as output. In K-Means and similar other algorithms, the amount of clusters, the initial data points have to be selected [9][10][11]. But in affinity propagation all data points are selected as potential exemplars, a real number *s(k,k)* is taken as input for each data point *k* that is known as "preference". Between the pairs of data points, messages are exchanged until high-quality clusters gradually emerge, containing each data point associated with a cluster. The preference values and the message-passing procedure have a great effect on the number of clusters. Preference value is specified as either 50 or medium of the input similarities. It also supports to take different kinds of a matrix as input where the shape of matrices will be [n_samples, n_features] or [n_samples, n_samples]. Performance comparison on K-Means and Affinity propagation is shown (Table 2) below that helps to realize why Affinity propagation is more applicable and acceptable nowadays.

Table 2. Comparison of the clustering methods in scikit-learn[3]

| Method name | Parameters | Scalability | Use Case |
| --- | --- | --- | --- |
| K-Means | Number of clusters | Large n_samples, Medium n_clusters. | Even cluster size, not too many clusters. |
| Affinity propagation | Damping, Preference | Not scalable with n_samples | Uneven cluster size, many clusters. |

---

[3] **https://scikit-learn.org/stable/modules/clustering.html**

We experimented on affinity propagation with coefficient similarity and median similarity using the sklearn library[4].

**Affinity Propagation With Coefficient Similarity:** In coefficient similarity, we used dice coefficient concept (section 4.2) between pairs of words based on shared unique di-grams and calculation is stored in a matrix with size [n_samples, n_samples] to fit the train data that is taken as input in the main function. In shared unique di-grams model not only *bi-grams* but *tri-grams* are also considered. The parameters of the main function: damping factor is set to 0.5, negative squared Euclidean distance between data points is considered as affinity value, the median of the input similarities is taken as preference value by default.

**Affinity Propagation With Median Similarity:** In median similarity, median distance is calculated between pairs of words and calculation is stored using a matrix with size [n_samples, n_samples] to fit the train data that is taken as input in the main function. One important point is if the median distance is larger than the minimum length of word pairs, it is separated by using a big value(i.e., 200). Parameters are same as coefficient similarity described earlier.

**5.0 Results And Evaluation**

Table 3. Detailed results of affinity propagation

| Affinity Propagation (Coefficient Similarity) | | Affinity Propagation (Median Similarity) | |
|---|---|---|---|
| Unique Token | 5540 | Unique Token | 5540 |
| Total Cluster | 488 | Total Cluster | 1305 |
| Correct clusters | 123 | Correct clusters | 650 |
| Correct word | 775 | Correct word | 1021 |

Using affinity propagation (Table 3), it is not possible to experiment with a large dataset. Because to fit the train data, affinity propagation creates a matrix with size [n_samples, n_samples]. When the number of tokens increases the matrix size will be increased. Storing and updating matrices of 'affinities', 'responsibilities' and 'similarities' between data points provides MemoryError on most machines. And by using affinity propagation we are failed to experiment over 5500 tokens both in coefficient similarity and median similarity. Although median similarity provides a large number of clusters, the number of correct clusters are failed to provide a satisfactory result in both cases.

Table 4. Detailed statistics of N-gram stemmer

| Topic | Quantity |
|---|---|
| Unique Token | 10133 |
| Total Cluster | 5972 |
| Correct Cluster | 5238 |
| Correct word | 8746 |
| Accuracy | 87% |

N-gram stemmer (Table 4) is capable of supporting large dataset. Our preprocessed dataset contains a total of 10133 unique words. It helps to reduce 10133 unique tokens to 5972 clustering root that promisingly will improve the retrieval effectiveness not only by saving space and time but also by increasing performance. From 5972 clusters 5238 clusters are accurate containing 8746 Correct words which mean 87% clustering root are accurate with satisfactory result. This experimental result shows its strength and efficiency by providing meaningful root word.

---

[4] **https://scikit-learn.org/stable/modules/generated/sklearn.cluster.AffinityPropagation.html#sklearn.cluster.AffinityPropagation**

**6.0 Conclusion And Future Direction**

In this paper, we proposed a Bangla stemmer by comparing an efficient affinity propagation clustering algorithm. It uses dice coefficient as an association measure to calculate the similarity between pairs of words based on shared unique di-gram method. By observing evaluation performed on N-gram stemmer it can be said that the stemmer has the ability to successfully cluster the lexically related sets of words and to create the corresponding lexical root word. It reduces 10133 unique words to 5972 clustering roots where 5238 roots are closely appropriate with accuracy 87% and appears as one of the plausible methods for enhancing Information Retrieval performance (IR) by providing meaningful root word.

We claimed that lexically related words should be in one cluster, but the system provides sub-clusters that is created because of word's length. Basically, when the length difference is larger than 3 in a lexically related word pair, the value of coefficient similarity provides a smaller result than the threshold value. At that time these two words are not placed in the same cluster. Although our subclusters are correct, we have to count them inappropriate. In future, we will try to solve this sub-clustering problem by using a more efficient method to improve the performance. Moreover, we have a plan to solve this problem by using a deep learning approach. Besides we will try to evaluate the n-gram stemming by using different similarity coefficients over a larger collection of data.